\documentclass[conference]{IEEEtran}
\usepackage{blindtext}
\ifCLASSINFOpdf
\else
\fi
\usepackage{verbatim}
\usepackage[pdftex]{graphicx,color}
\graphicspath{{figure/}}
\DeclareGraphicsExtensions{.pdf,.jpeg,.png}

\usepackage{algorithmicx}
\usepackage{algpseudocode}
\usepackage{algorithm}
\usepackage{amssymb}
\usepackage{amsmath}
\usepackage{amsfonts}
\usepackage{amsthm}
\usepackage{colortbl}
\usepackage{tablefootnote}
\usepackage{subcaption}

\newcommand{\mb}{\mathbf}
\newcommand{\mc}{\mathcal}
\newcommand{\mbb}{\mathbb}
\newcommand{\E}{\left\vert{\mc{E}}\right\vert}

\newcommand{\StartCompact}[1]{\par\vspace{-0.75em}\begin{#1}\allowdisplaybreaks\ignorespaces}
\newcommand{\StopCompact}[1]{\end{#1}\ignorespaces}
\newtheorem{theorem}{Theorem}
\newtheorem{corollary}{Corollary}

\newtheorem{lemma}{Lemma}

\newenvironment{myproof}{\noindent\textbf{Proof:}}{\hfill$\square$}

\begin{document}
%
\title{Online Learning for Wireless Distributed Computing}

\author{\IEEEauthorblockN{Yi-Hsuan Kao\IEEEauthorrefmark{1},
Kwame Wright\IEEEauthorrefmark{1},
Bhaskar Krishnamachari\IEEEauthorrefmark{1} and
Fan Bai\IEEEauthorrefmark{2}}
\IEEEauthorblockA{\IEEEauthorrefmark{1}Ming-Hsieh Department of Electrical Engineering\\
University of Southern California,
Los Angeles, CA, USA\\ Email: \{yihsuank, kwamelaw, bkrishna\}@usc.edu}
\IEEEauthorblockA{\IEEEauthorrefmark{2}General Motors Global R\&D\\
Warren, MI, USA\\
Email: fan.bai@gm.com}
}


%


\maketitle

\begin{abstract}
There has been a growing interest for Wireless Distributed Computing (WDC), which leverages collaborative computing over multiple wireless devices. WDC enables complex applications that a single device cannot support individually. However, the problem of assigning tasks over multiple devices becomes challenging in the dynamic environments encountered in real-world settings, considering that the resource availability and channel conditions change over time in unpredictable ways due to mobility and other factors. In this paper, we formulate a task assignment problem as an online learning problem using an adversarial multi-armed bandit framework. We propose MABSTA, a novel online learning algorithm that learns the performance of unknown devices and channel qualities continually through exploratory probing and makes task assignment decisions by exploiting the gained knowledge. For maximal adaptability, MABSTA is designed to make no stochastic assumption about the environment. We analyze it mathematically and provide a worst-case performance guarantee for any dynamic environment. We also compare it with the optimal offline policy as well as other baselines via emulations on trace-data obtained from a wireless IoT testbed, and show that it offers competitive and robust performance in all cases. To the best of our knowledge, MABSTA is the first online algorithm in this domain of task assignment problems and provides provable performance guarantee.
\end{abstract}

%
\IEEEpeerreviewmaketitle

\section{Introduction}
We are at the cusp of revolution as the number of connected devices is projected to grow significantly in the near future. These devices, either suffering from stringent battery usage, like mobile devices, or limited processing power, like sensors, are not capable to run computation-intensive tasks independently. Nevertheless, what can these devices do if they are connected and collaborate with each other? The connected devices in the network, sharing resources with each other, provide a platform with abundant computational resources that enables the execution of complex applications \cite{datla2012wireless,arslancwc}.

Traditional cloud services provide access to high performance and reliable servers. However, considering the varying link quality and the long run trip times (RTTs) of a wide-area network (WAN) and possibly long setup time, these remote servers might not always be the best candidates to help in scenarios where the access delay is significant \cite{shi2014cosmos, li2010cloudcmp}. Another approach is to exploit nearby computational resources, including mobile devices, road-side units (RSUs) and local servers. These devices are not as powerful as cloud servers in general, but can be accessed by faster device to device (D2D) communication \cite{shi2012serendipity}. In addition to communication over varying wireless links, the workload on a device also affects the amount of resource it can release. Hence, a system has to identify the available resources in the network and decide how to leverage them among a number of possibilities, considering the dynamic environment at run time.

\begin{figure}
\centering
\includegraphics[scale = 0.32]{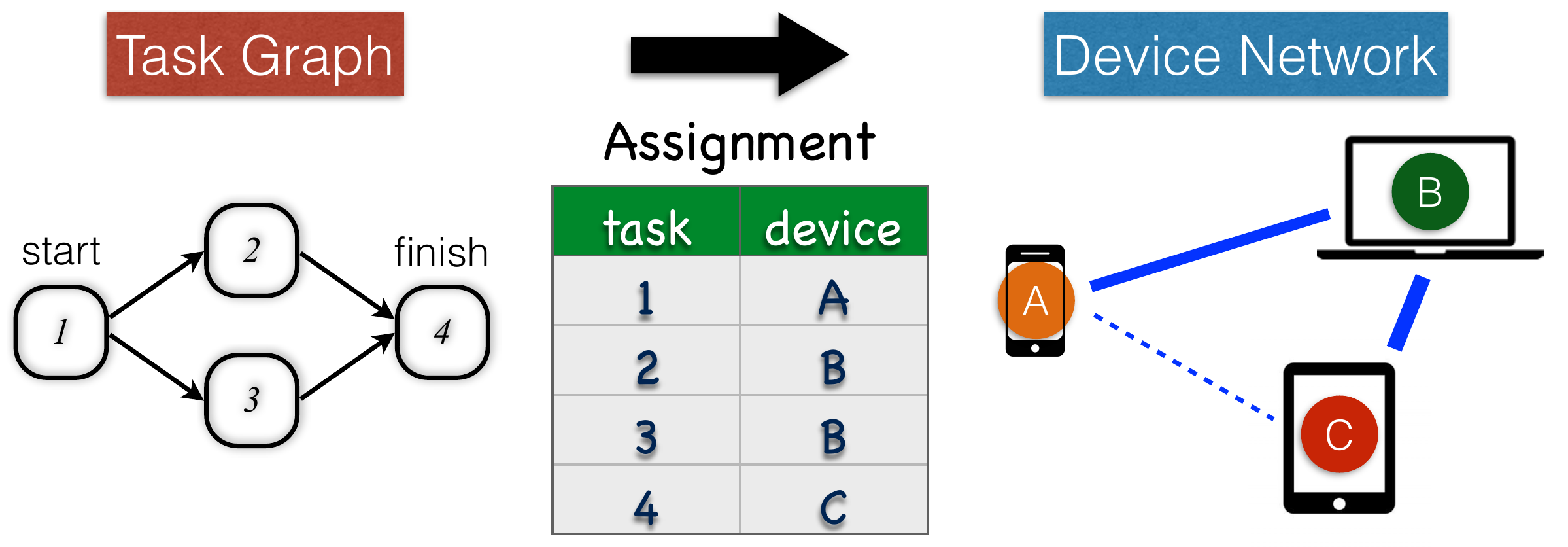}
\caption{An application consists of multiple tasks. In order to perform collaborative computing over heterogeneous devices connected in the network, a system has to find out a good task assignment strategy, considering devices' feature, workload and different channel qualities between them.}
\label{fig: intro}
\end{figure}

Figure \ref{fig: intro} illustrates the idea of Wireless Distributed Computing. Given an application that consists of multiple tasks, we want to assign them on multiple devices, considering the resource availability so that the system performance, in metrics like energy consumption and application latency, can be improved. These resources that are accessible by wireless connections form a resource network, which is subject to frequent topology changes and has the following features:

\noindent \textbf{Dynamic device behavior}: The quantity of the released resource varies with devices, and may also depend on the local processes that are running. Moreover, some of devices may carry microporcessors that are specialized in performing a subset of tasks. Hence, the performance of each device varies highly over time and different tasks and is hard to model as a known and stationary stochastic process.

\noindent \textbf{Heterogeneous network with intermittent connections}: Devices' mobility makes the connections intermittent, which change drastically in quality within a short time period. Furthermore, different devices may use different protocols to communicate with each other. Hence, the performance of the links between devices is also highly dynamic and variable and hard to model as a stationary process.

\subsection{Why online learning?}
From what we discuss above, since the resource network is subject to drastic changes over time and is hard to be modeled by stationary stochastic processes, we need an algorithm that applies to \textit{all} possible scenarios, learns the environment at run time, and adapts to changes. Existing works focus on solving optimization problems given known deterministic profile or known stochastic distributions \cite{chen2010cross, kaohermes}. These problems are hard to solve. More importantly, algorithms that lack learning ability could be harmed badly by statistical changes or mismatch between the profile (offline training) and the run-time environment. Hence, we use an online learning approach, which takes into account the performance during the learning phase, and aim to learn the environment quickly and adapt to changes.

We formulate the task assignment problem as an adversarial multi-armed bandit (MAB) problem that does not make any stochastic assumptions on the resource network \cite{auer1995gambling}. We propose MABSTA (Multi-Armed Bandit based Systematic Task Assignment) that learns the environment and makes task assignment at run time. Furthermore, We provide worst-case analysis on the performance to guarantee that MABSTA performs no worse than a provable lower bound in \textit{any} dynamic environment. To the best of our knowledge, MABSTA is the first online algorithm in this domain of task assignment problems and provides provable performance guarantee.

\subsection{Contributions}
\noindent\textbf{A new formulation of task assignment problems considering general and dynamic environment: } We use a novel adversarial multi-armed bandit (MAB) formulation that does not make any assumptions on the dynamic environment. That is, it applies to all realistic scenarios.

\noindent\textbf{A light algorithm that learns the environment quickly with provable performance guarantee: } MABSTA runs with light complexity and storage, and admits performance guarantee and learning time that are significantly improved compared to the existing MAB algorithm.

\noindent\textbf{Broad applications on wireless device networks:} MABSTA enhances collaborative computing over wireless devices, enabling more potential applications on mobile cloud computing, wireless sensor networks and Internet of Things. 

\section{Background on Multi-armed Bandit Problems}
\label{sec: background}
The multi-armed bandit (MAB) problem is a sequential decision problem where at each time an agent chooses over a set of ``arms", gets the payoff from the selected arms and tries to learn the statistical information from sensing them. These formulations have been considered recently in the context of opportunistic spectrum access for cognitive radio wireless networks, but those formulations are quite different from ours in that they focus only on channel allocation and not on also allocating computational tasks to servers \cite{dai2014online,liu2010indexability}.

Given an online algorithm to a MAB problem, its performance is measured by a regret function, which specifies how much the agent loses due to the unknown information at the beginning \cite{bubeck2012regret}. For example, we can compare the performance to a genie who knows the statistics of payoff functions and selects the arms based on the best policy.

Stochastic MAB problems model the payoff of each arm as a stationary random process and aim to learn the unknown information behind it. If the distribution is unknown but is known to be i.i.d. over time, Auer \emph{et al.} \cite{auer2002finite} propose UCB algorithms to learn the unknown distribution with bounded regret. However, the assumption on i.i.d. processes does not always apply to the real environment. On the other hand, Ortner \emph{et al.} \cite{ortner2012regret} assume the distribution is known to be a Markov process and propose an algorithm to learn the unknown state transition probabilities. However, the large state space of Markov process causes our task assignment problem to be intractable. Hence, we need a \textit{tractable} algorithm that applies to stochastic processes with \textit{relaxed} assumptions on time-independence stationarity.

Adversarial MAB problems, however, do not make any assumptions on the payoffs. Instead, an agent learns from the sequence given by an adversary who has complete control over the payoffs \cite{auer1995gambling}. In addition to the well-behaved stochastic processes, an algorithm of adversarial MAB problems gives a solution that generally applies to all bounded payoff sequences and provides the the worst-case performance guarantee.

Auer \emph{et al.} \cite{auer2002nonstochastic} propose Exp3, which serves adversarial MAB and yields a sub-linear regret with time ($O(\sqrt{T})$). That is, compared to the optimal offline algorithm, Exp3 achieves asymptotically $1$-competitive. However, if we apply Exp3 to our task assignment problem, there will be an exponential number of arms, hence, the regret will grow exponentially with problem size. In this paper, we propose an algorithm providing that the regret is not only bounded by $O(\sqrt{T})$ but also bounded by a polynomial function of problem size.

\section{Problem Formulation}

Suppose a data processing application consists of $N$ tasks, where their dependencies are described by a directed acyclic graph (DAG) $G = (\mc{V},\mc{E})$ as shown in Figure \ref{fig: intro}. That is, an edge $(m,n)$ implies that some data exchange is necessary between task $m$ and task $n$ and hence task $n$ cannot start until task $m$ finishes. There is an incoming data stream to be processed ($T$ data frames in total), where for each data frame $t$, it is required to go through all the tasks and leave afterwords. There are $M$ available devices. The assignment strategy of data frame $t$ is denoted by a vector $\mb{x}^t = {x^t_1, \cdots, x^t_N}$, where $x^t_i$ denotes the device that executes task $i$. Given an assignment strategy, stage-wised costs apply to each node (task) for computation and each edge for communication. The cost can correspond to the resource consumption for a device to complete a task, for example, energy consumption.

In the following formulation we follow the tradition in MAB literature and focus on maximizing a positive reward instead of minimizing the total cost, but of course these are mathematically equivalent, e.g., by setting $reward = maxCost - cost$. When processing data frame $t$, let $R_i^{(j)}(t)$ be the reward of executing task $i$ on device $j$. Let $R_{mn}^{(jk)}(t)$ be the reward of transmitting the data of edge $(m,n)$ from device $j$ to $k$. The reward sequences are unknown but are bounded between $0$ and $1$. Our goal is to find out the assignment strategy for each data frame based on the previously observed samples, and compare the performance with a genie that uses the best assignment strategy for all data frames. That is,

\StartCompact{small}
\begin{equation}
R_{total}^{max} = \max_{\mb{x} \in \mc{F}} \sum_{t=1}^T \left(\sum_{i=1}^N R_{i}^{(x_i)}(t) + \sum_{(m,n) \in \mc{E}} R_{mn}^{(x_mx_n)}(t)\right),
\label{eq:formulation}
\end{equation}
\StopCompact{small}
where $\mc{F}$ represents the set of feasible solutions. The genie who knows all the reward sequences can find out the best assignment strategy, however, not knowing these sequences in advance, our proposed online algorithm aims to learn this best strategy and remain competitive in overall performance.

\section{MABSTA Algorithm}

\begin{algorithm}[t]
\caption{MABSTA}\label{alg: MABSTA}
\begin{algorithmic}[1]
\Procedure{MABSTA}{$\gamma,\alpha$}
\State $w_{\mb{y}}(1) \gets 1 \; \forall \mb{y} \in \mc{F}$
\For{$t\gets 1,2,\cdots,T$}
\State $W_t \gets \sum_{\mb{y} \in \mc{F}}w_{\mb{y}}(t)$
\State Draw $\mb{x}^t$ from distribution
\begin{equation}
p_{\mb{y}}(t) = (1-\gamma)\frac{w_{\mb{y}}(t)}{W_t} + \frac{\gamma}{\left\vert\mc{F}\right\vert}
\label{eq: prob}
\end{equation}
\State Get rewards $\{R_{i}^{(x^t_i)}(t)\}_{i=1}^N$, $\{R_{mn}^{(x^t_mx^t_n)}(t)\}_{(m,n) \in \mc{E}}$.
\State $\mc{C}_{ex}^i \gets \{\mb{z} \in \mc{F}|z_i = x^t_i\}, \; \forall i$
\State $\mc{C}_{tx}^{mn} \gets \{\mb{z} \in \mc{F}|z_m = x^t_m,z_n = x^t_n\}, \; \forall (m,n)$
\For{$\forall j \in [M]$, $\forall i \in [N]$}
\StartCompact{small}
\begin{equation}
\hat{R}_{i}^{(j)}(t) = 
\begin{cases}
\frac{R_{i}^{(j)}(t)}{\sum_{\mb{z} \in \mc{C}_{ex}^i}p_{\mb{z}}(t)} & \; \text{if } x^t_i = j, \\
0 & \; \text{otherwise.}
\end{cases}
\label{eq: updateEstimateNode}
\end{equation}
\StopCompact{small}
\EndFor
\For{$\forall j,k \in [M]$, $\forall (m,n) \in \mc{E}$}
\StartCompact{small}
\begin{equation}
\hat{R}_{mn}^{(jk)}(t) = 
\begin{cases}
\displaystyle \frac{R_{mn}^{(jk)}(t)}{\sum_{\mb{z} \in \mc{C}_{tx}^{mn}}p_{\mb{z}}(t)} & \; \text{if } x^t_m = j, x^t_n = k, \\
0 & \; \text{otherwise.}
\end{cases}
\label{eq: updateEstimateEdge}
\end{equation}
\StopCompact{small}
\EndFor
\State Update for all $\mb{y}$
\StartCompact{small}
\begin{equation}
\hat{R}_{\mb{y}}(t) = \sum_{i=1}^N \hat{R}_{i}^{(y_i)}(t) + \sum_{(m,n) \in \mc{E}} \hat{R}_{mn}^{(y_my_n)}(t),
\label{eq: estimate}
\end{equation}
\begin{equation}
w_{\mb{y}}(t+1) = w_{\mb{y}}(t)\exp\left(\alpha \hat{R}_{\mb{y}}(t)\right).
\label{eq: update}
\end{equation}
\StopCompact{small}
\EndFor
\EndProcedure
\end{algorithmic}
\end{algorithm}

We summarize MABSTA in Algorithm \ref{alg: MABSTA}. For each data frame $t$, MABSTA randomly selects a feasible assignment (arm $\mb{x} \in \mc{F}$) from a probability distribution that depends on the weights of arms ($w_{\mb{y}}(t)$). Then it updates the weights based on the reward samples. From (\ref{eq: prob}), MABSTA randomly switches between two phases: exploitation (with probability $1-\gamma$) and exploration (with probability $\gamma$). At exploitation phase, MABSTA selects an arm based on its weight. Hence, the one with higher reward samples will be chosen more likely. At exploration phase, MABSTA uniformly selects an arm without considering its performance. The fact that MABSTA keeps probing every arms makes it adaptive to the changes of the environment, compared to the case where static strategy plays the previously best arm all the time without knowing that other arms might have performed better currently.

The commonly used performance measure for an MAB algorithm is its regret. In our case it is defined as the difference in accumulated rewards ($\hat{R}_{total}$) compared to a genie that knows all the rewards and selects a single best strategy for all data frames ($R_{total}^{max}$ in (\ref{eq:formulation})). Auer \emph{et al.} \cite{auer2002nonstochastic} propose Exp3 for adversarial MAB. However, if we apply Exp3 to our online task assignment problem, since we have an exponential number of arms ($M^N$), the regret bound will grow exponentially. The following theorem shows that MABSTA guarantees a regret bound that is polynomial with problem size and $O(\sqrt{T})$. 
\begin{theorem}
\label{th:1}
Assume all the reward sequences are bounded between $0$ and $1$. Let $\hat{R}_{total}$ be the total reward achieved by Algorithm \ref{alg: MABSTA}. For any $\gamma \in (0,1)$, let $\alpha = \frac{\gamma}{M(N+\E M)}$, we have
\StartCompact{small}
\begin{equation*}
R_{total}^{max} - \mbb{E}\{\hat{R}_{total}\} \leq (e-1)\gamma R_{total}^{max} + \frac{M(N+\left\vert{\mc{E}}\right\vert M) \ln M^N}{\gamma}.
\end{equation*}
\StopCompact{small}
\end{theorem}
In above, $N$ is the number of nodes (tasks) and $\E$ is the number of edges in the task graph. We leave the proof of Theorem \ref{th:1} in the appendix. By applying the appropriate value of $\gamma$ and using the upper bound $R_{total}^{max} \leq (N+\E)T$, we have the following Corollary.
\begin{corollary}
Let $\gamma = \min \{1,\sqrt{\frac{M(N+\left\vert{\mc{E}}\right\vert M) \ln M^N}{(e-1)(N+\left\vert{\mc{E}}\right\vert)T}}\}$, then
\StartCompact{small}
\begin{equation*}
R_{total}^{max} - \mbb{E}\{\hat{R}_{total}\} \leq 2.63 \sqrt{(N+\E)(N+\E M)MNT \ln M}.
\end{equation*}
\StopCompact{small}
\label{col:1}
\end{corollary}
We look at the worst case, where $\E = O(N^2)$. The regret can be bounded by $O(N^{2.5}MT^{0.5})$. Since the bound is a concave function of $T$, we define the learning time $T_0$ as the time when its slope falls below a constant $c$. That is,
\begin{equation*}
T_0 = \frac{1.73}{c^2}(N+\E)(N+\E M)MN \ln M.
\end{equation*}
This learning time is significantly improved compared with applying Exp3 to our problem, where $T_0 = O(M^N)$. As we will show in the numerical results, MABSTA performs significantly better than Exp3 in the trace-data emulation.

\section{Polynomial Time MABSTA}
In Algorithm \ref{alg: MABSTA}, since there are exponentially many arms, implementation may result in exponential storage and complexity. However, in the following, we propose an equivalent but efficient implementation. We show that when the task graph belongs to a subset of DAG that appear in practical applications (namely, parallel chains of trees), Algorithm \ref{alg: MABSTA} can run in polynomial time with polynomial storage.

We observe that in (\ref{eq: estimate}), $R_{\mb{y}}(t)$ relies on the estimates of each node and each edge. Hence, we rewrite (\ref{eq: update}) as
\StartCompact{small}
\begin{align}
\notag w_{\mb{y}}(t+1) &= \exp \left(\alpha \sum_{\tau = 1}^t R_{\mb{y}}(t)\right) \\ 
&= \exp \left(\alpha \sum_{i=1}^N \tilde{R}_i^{(y_i)}(t) + \alpha\sum_{(m,n) \in \mc{E}} \tilde{R}_{mn}^{(y_my_n)}(t)\right),
\label{eq: decompose}
\end{align}
\StopCompact{small}
where
\begin{equation*}
\tilde{R}_i^{(y_i)}(t) = \sum_{\tau = 1}^t \hat{R}_i^{(y_i)},\;\;
\tilde{R}_{mn}^{(y_my_n)}(t) = \sum_{\tau = 1}^t \hat{R}_{mn}^{(y_my_n)}.
\end{equation*}
To calculate $w_{\mb{y}}(t)$, it suffices to store $\tilde{R}_i^{(j)}(t)$ and $\tilde{R}_{mn}^{(j,k)}(t)$ for all $i \in [N]$, $(m,n) \in \mc{E}$ and $j,k \in [M]$, which cost $(NM + \E M^2)$ storage.

Equation (\ref{eq: updateEstimateNode}) and (\ref{eq: updateEstimateEdge}) require the knowledge of marginal probabilities $\mbb{P}\{x^t_i = j\}$ and $\mbb{P}\{x^t_m = j, x^t_n = k\}$. Next, we propose a polynomial time algorithm to calculate them. From (\ref{eq: prob}), the marginal probability can be written as
\[
\mbb{P}\{x^t_i = j\} = (1-\gamma)\frac{1}{W_t}\sum_{\mb{y}:y_i = j}w_{\mb{y}}(t) + \frac{\gamma}{M}.
\]
Hence, without calculating $W_t$, we have
\StartCompact{small}
\begin{equation}
\mbb{P}\{x^t_i = j\}-\frac{\gamma}{M}:\mbb{P}\{x^t_i = k\}-\frac{\gamma}{M} = \sum_{\mb{y}:y_i = j}w_{\mb{y}}(t) : \sum_{\mb{y}:y_i = k}w_{\mb{y}}(t).
\label{eq: ratio}
\end{equation}
\StopCompact{small}

\subsection{Tree-structure Task Graph}

\begin{algorithm}[t]
\caption{Calculate $w_N^{(j)}$ for tree-structured task graph}\label{alg: tree}
\begin{algorithmic}[1]
\Procedure{$\Omega$}{$N,M,G$}
\State $q \gets \text{BFS}\left(G,N\right)$\Comment{run BFS from $N$ and store visited nodes in order}
\For{$i\gets q\text{.end}, \; q\text{.start}$}\Comment{start from the last element}
\If{$i$ is a leaf}\Comment{initialize $\omega$ values of leaves}
\State
\[
\omega_i^{(j)} \gets e_i^{(j)}
\]
\Else
\State
\[
\omega_i^{(j)} \gets e_i^{(j)}\prod_{m \in \mc{N}_i}\sum_{y_m \in [M]} e_{mi}^{(y_mj)}\omega_m^{(y_m)}
\]
\EndIf
\EndFor
\EndProcedure
\end{algorithmic}
\end{algorithm}

\begin{figure}[t]
\centering
\includegraphics[scale = 0.4]{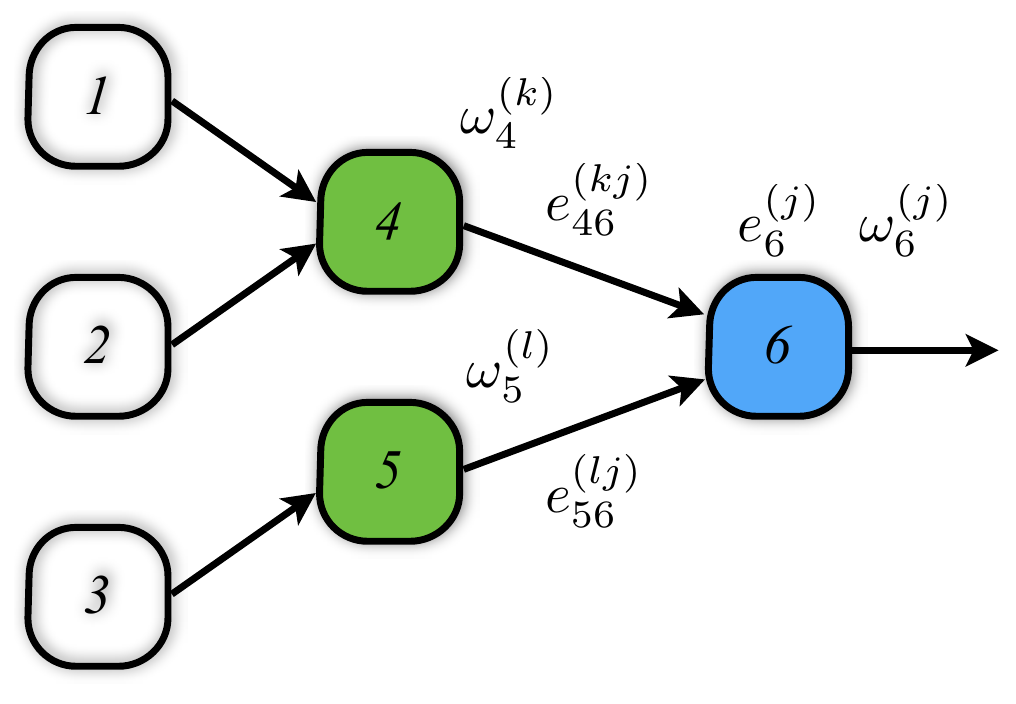}
\caption{An example of tree-structure task graph, where $\mc{D}_6 = \{1,2,3,4,5\}$, and $\mc{E}_6 = \{(1,4),(2,4),(3,5),(4,6),(5,6)\}$.}
\label{fig: appTree}
\end{figure}

Now we focus on how to calculate the sum of weights in (\ref{eq: ratio}) efficiently. We start from tree-structure task graphs and solve the more general graphs by calling the proposed algorithm for trees a polynomial number of times.

We drop time index $t$ in our derivation whenever the result holds for all time steps $t \in \{1,\cdots,T\}$. For example, $\tilde{R}_i^{(j)} \equiv \tilde{R}_i^{(j)}(t)$. We assume that the task graph is a tree with $N$ nodes where the $N^\textrm{th}$ node is the root (final task). Let $e_i^{(j)} = \exp(\alpha \tilde{R}_i^{(j)})$ and $e_{mn}^{(jk)} = \exp(\alpha \tilde{R}_{mn}^{(jk)})$. Hence, the sum of exponents in (\ref{eq: decompose}) can be written as the product of $e_i^{(j)}$ and $e_{mn}^{(jk)}$. That is,
\begin{equation*}
\sum_{\mb{y}}w_{\mb{y}}(t) = \sum_{\mb{y}}\prod_{i=1}^N e_i^{(y_i)}\prod_{(m,n) \in \mc{E}} e_{mn}^{(y_my_n)}.
\end{equation*}

For a node $v$, we use $\mc{D}_v$ to denote the set of its descendants. Let the set $\mc{E}_v$ denote the edges connecting its descendants. Formally,
\[
\mc{E}_v = \{(m,n) \in \mc{E}|m \in \mc{D}_v, n \in \mc{D}_v \cup \{v\}\}.
\]
The set of $\left\vert{\mc{D}_v}\right\vert$-dimensional vectors, $\{y_m\}_{m \in \mc{D}_v}$, denotes all the possible assignments on its descendants. Finally, we define the sub-problem, $\omega_i^{(j)}$, which calculates the sum of weights of all possible assignment on task $i$'s descendants, given task $i$ is assigned to device $j$. That is,
\begin{equation}
\omega_i^{(j)} = e_i^{(j)}\sum_{\{y_m\}_{m \in \mc{D}_i}}\prod_{m \in \mc{D}_i}e_m^{(y_m)}\prod_{(m,n) \in \mc{E}_i}e_{mn}^{(y_my_n)}.
\label{eq: omega}
\end{equation}

Figure \ref{fig: appTree} shows an example of a tree-structure task graph. Task $4$ and $5$ are the children of task $6$. From (\ref{eq: omega}), if we have $\omega_4^{(k)}$ and $\omega_5^{(l)}$ for all $k$ and $l$, $\omega_6^{(j)}$ can be solved by
\[
\omega_6^{(j)} = e_6^{(j)}\sum_{k,l}e_{46}^{(kj)}\omega_4^{(k)}e_{56}^{(lj)}\omega_5^{(l)}.
\]
In general, the relation of weights between task $i$ and its children $m \in \mc{N}_i$ is given by the following equation.

\begin{align}
\notag \omega_i^{(j)} &= e_i^{(j)}\sum_{\{y_m\}_{m \in \mc{N}_i}}\prod_{m \in \mc{N}_i}e_{mi}^{(y_mj)}\omega_m^{(y_m)} \\
&= e_i^{(j)}\prod_{m \in \mc{N}_i}\sum_{y_m \in [M]} e_{mi}^{(y_mj)}\omega_m^{(y_m)}.
\end{align}

Algorithm \ref{alg: tree} summarizes our approach to calculate the sum of weights of a tree-structure task graph. We first run breath first search (BFS) from the root node. Then we start solving the sub-problems from the last visited node such that when solving task $i$, it is guaranteed that all of its child tasks have been solved. Let $d_{in}$ denote the maximum in-degree of $G$ (i.e., the maximum number of in-coming edges of a node). Running BFS takes polynomial time. For each sub-problem, there are at most $d_{in}$ products of summations over $M$ terms. In total, Algorithm \ref{alg: tree} solves $NM$ sub-problems. Hence, Algorithm \ref{alg: tree} runs in $\Theta(d_{in}NM^2)$ time.

\subsection{More general task graphs}

\begin{figure}[t]
\centering
\includegraphics[scale = 0.4]{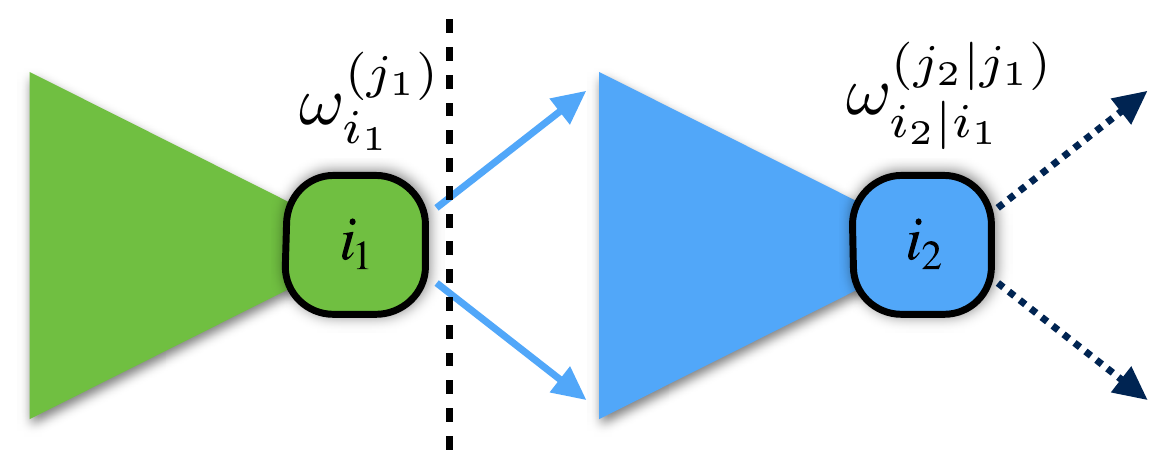}
\caption{A task graph consists of serial trees. To solve the sum of weights, $\omega_{i_2}^{(j_2)}$, we solve two trees rooted from $i_1$ and $i_2$ separately. When solving $i_2$, we solve the conditional cases on all possible assignments of node $i_1$.}
\label{fig: appSerialTree}
\end{figure}

All of the nodes in a tree-structure task graph have only one out-going edge. For task graphs where there exists a node that has multiple out-going edges, we decompose the task graph into multiple trees and solve them separately and combine the solutions in the end. In the following, we use an example of a task graph that consists of serial trees to illustrate our approach.

Figure \ref{fig: appSerialTree} shows a task graph that has two trees rooted by task $i_1$ and $i_2$, respectively. Let the sub-problem, $\omega_{i_2|i_1}^{(j_2|j_1)}$, denote the sum of weights given that $i_2$ is assigned to $j_2$ and $i_1$ is assigned to $j_1$. To find $\omega_{i_2|i_1}^{(j_2|j_1)}$, we follow Algorithm \ref{alg: tree} but consider the assignment on task $i_1$ when solving the sub-problems on each leaf $m$. That is,
\[
\omega_{(m|i_1)}^{j_m|j_1} = e_{i_1m}^{(j_1j_m)}e_m^{(j_m)}.
\]
The sub-problem, $\omega_{i_2}^{(j_2)}$, now becomes the sum of weights of all possible assignment on task $i_2$'s descendants, including task $1$'s descendants, and is given by
\begin{equation}
\omega_{i_2}^{(j_2)} = \sum_{j_1 \in [M]}w_{i_2|i_1}^{(j_2|j_1)}w_{i_1}^{(j_1)}.
\label{eq: combine}
\end{equation}
For a task graph that consists of serial trees rooted by $i_1,\cdots,i_n$ in order, we can solve $\omega_{i_r}^{(j_r)}$, given previously solved $\omega_{i_r|i_{r-1}}^{(j_r|j_{r-1})}$ and $\omega_{i_{r-1}}^{(j_{r-1})}$. From (\ref{eq: combine}), to solve $\omega_{i_2}^{(j_2)}$, we have to solve $w_{i_2|i_1}^{(j_2|j_1)}$ for $j_1 \in \{1,\cdots,M\}$. Hence, it takes $O(d_{in}n_1M^2) + O(Md_{in}n_2M^2)$ time, where $n_1$ (resp. $n_2$) is the number of nodes in tree $i_1$ (resp. $i_2$). Hence, to solve a serial-tree task graph, it takes $O(d_{in}NM^3)$ time.

Our approach can be generalized to more complicated DAGs, like the one that contains parallel chains of trees (parallel connection of Figure \ref{fig: appSerialTree}), in which we solve each chain independently and combine them from their common root $N$. Most of the real applications can be described by these families of DAGs where we have proposed polynomial time MABSTA to solve them. For example, in \cite{ra2011odessa}, the three benchmarks fall in the category of parallel chains of trees. In Wireless Sensor Networks, an application typically has a tree-structured workflow \cite{viswanathan2013enabling}.

\subsection{Marginal Probability}
From (\ref{eq: ratio}), we can calculate the marginal probability $\mbb{P}\{x^t_i = j\}$ if we can solve the sum of weights over all possible assignments given task $i$ is assigned to device $j$. If task $i$ is the root (node $N$), then Algorithm \ref{alg: tree} solves $\omega_{i}^{(j)} = \sum_{\mb{y}:y_i = j}w_{\mb{y}}(t)$ exactly. If task $i$ is not the root, we can still run Algorithm \ref{alg: tree} to solve $[\omega_p^{(j^\prime)}]_{y_i = j}$, which fixes the assignment of task $i$ to device $j$ when solving from $i$'s parent $p$. That is,
\[
[\omega_p^{(j^\prime)}]_{y_i = j} = e_p^{(j^\prime)}e_{ip}^{(jj^\prime)}\omega_i^{(j)}\prod_{m \in \mc{N}_p \setminus \{i\}}\sum_{y_m}e_{mp}^{(y_mj^\prime)}\omega_m^{(y_m)}.
\]
Hence, in the end, we can solve $[\omega_N^{(j^\prime)}]_{y_i = j}$ from the root and 
\[
\sum_{\mb{y}:y_i = j}w_{\mb{y}}(t) = \sum_{j^\prime \in [M]}[\omega_r^{(j^\prime)}]_{y_i = j}.
\]
Similarly, the $\mbb{P}\{x^t_m = j, x^t_n = k\}$ can be achieved by solving the conditional sub-problems on both tasks $m$ and $n$.

\subsection{Sampling}

\begin{algorithm}[!t]
\caption{Efficient Sampling Algorithm}\label{alg: sampling}
\begin{algorithmic}[1]
\Procedure{Sampling}{$\gamma$}
\State $s \gets rand()$ \Comment{get a random number between $0$ and $1$}
\If {$s < \gamma$}
\State pick an $\mb{x} \in [M]^N$ uniformly
\Else
\For{$i \gets 1,\cdots,N$}
\State $[\omega_i^{(j)}]_{x^t_1,\cdots,x^t_{i-1}} \gets \Omega(N,M,G)_{x^t_1,\cdots,x^t_{i-1}}$
\State $\mbb{P}\{x^t_i = j|x^t_1,\cdots,x^t_{i-1}\} \propto [\omega_i^{(j)}]_{x^t_1,\cdots,x^t_{i-1}}$
\EndFor
\EndIf
\EndProcedure
\end{algorithmic}
\end{algorithm}

As we can calculate the marginal probabilities efficiently, we propose an efficient sampling policy summarized in Algorithm \ref{alg: sampling}. Algorithm \ref{alg: sampling} first selects a random number $s$ between $0$ and $1$. If $s$ is less than $\gamma$, it refers to the exploration phase, where MABSTA simply selects an arm uniformly. Otherwise, MABSTA selects an arm based on the probability distribution $p_{\mb{y}}(t)$, which can be written as
\begin{align*}
p_{\mb{y}}(t) &= \mbb{P}\{x^t_1 = y_1\} \cdot \mbb{P}\{x^t_2 = y_2 | x^t_1 = y_1\}\\
&\cdots\mbb{P}\{x^t_N = y_N | x^t_1 = y_1, \cdots, x^t_{N-1} = y_{N-1}\}.
\end{align*}
Hence, MABSTA assigns each task in order based on the conditional probability given the assignment on previous tasks. For each task $i$, the conditional probability can be calculate efficiently by running Algorithm \ref{alg: tree} with fixed assignment on task $1,\cdots,i-1$.

\section{Numerical Evaluation}
In this section, we first examine how MABSTA adapts to dynamic environment. Then, we perform trace-data emulation to verify MABSTA's performance guarantee and compare it with other algorithms.
\subsection{MABSTA's Adaptivity}
\begin{figure}
\centering
\includegraphics[scale = 0.3]{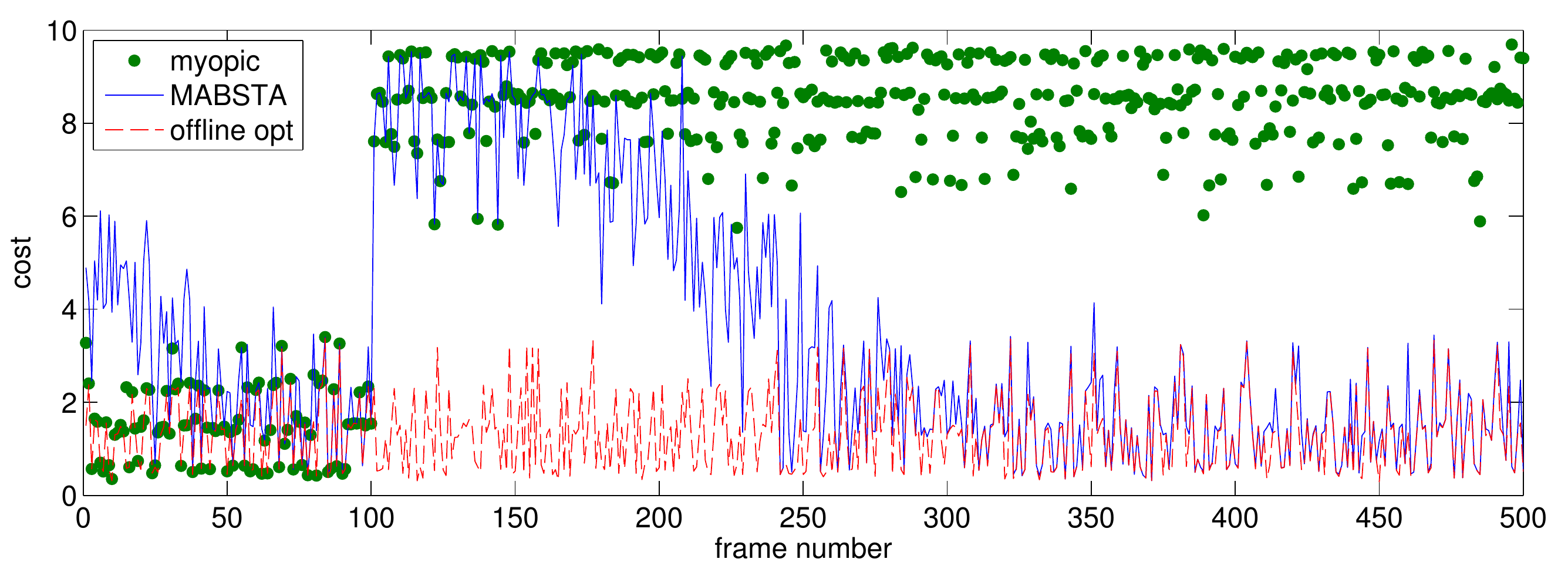}
\caption{MABSTA adapts to the changes at the $100^\textrm{th}$ frame, while the myopic policy, relying on the old information of the environment, fails to adjust the task assignment.}
\label{fig: adaptivity}
\end{figure}

Here we examine MABSTA's adaptivity to dynamic environment and compare it to the optimal strategy that relies on the existing profile. We use a two-device setup, where the task execution costs of the two devices are characterized by two different Markov processes. We neglect the channel communication cost so that the optimal strategy is the myopic strategy. That is, assigning the tasks to the device with the highest belief that it is in ``good" state \cite{dirickx1975optimality}. We run our experiment with an application that consists of $10$ tasks and processes the in-coming data frames one by one. The environment changes at the $100^\textrm{th}$ frame, where the transition matrices of two Markov processes swap with each other. From Figure \ref{fig: adaptivity}, there exists an optimal assignment (dashed line) so that the performance remains as good as it was before the $100^\textrm{th}$ frame. However, myopic strategy, with the wrong information of the transition matrices, fails to adapt to the changes. From (\ref{eq: prob}), MABSTA not only relies on the result of previous samples but also keeps exploring uniformly (with probability $\frac{\gamma}{M^N}$ for each arm). Hence, when the performance of one device degrades at $100^\textrm{th}$ frame, the randomness enables MABSTA to explore another device and learn the changes.

\subsection{Trace-data Emulation}

\begin{table}
\renewcommand{\arraystretch}{1.1}
\centering
\caption{Parameters Used in Trace-data measurement}
\label{table: parameter}
\begin{tabular}{|c|c||c|c|}
\hline
Device ID & \# of iterations & Device ID & \# of iterations\\
\hline
\hline
18 & $\mc{U}(14031, 32989)$ & 28 & $\mc{U}(10839, 58526)$\\
\hline
21 & $\mc{U}(37259, 54186)$ & 31 & $\mc{U}(10868, 28770)$\\
\hline
22 & $\mc{U}(23669, 65500)$ & 36 & $\mc{U}(41467, 64191)$\\
\hline
24 & $\mc{U}(61773, 65500)$ & 38 & $\mc{U}(12386, 27992)$\\
\hline
26 & $\mc{U}(19475, 44902)$ & 41 & $\mc{U}(15447, 32423)$\\
\hline\end{tabular}
\end{table}

\begin{figure}
\centering
\includegraphics[scale = 0.42]{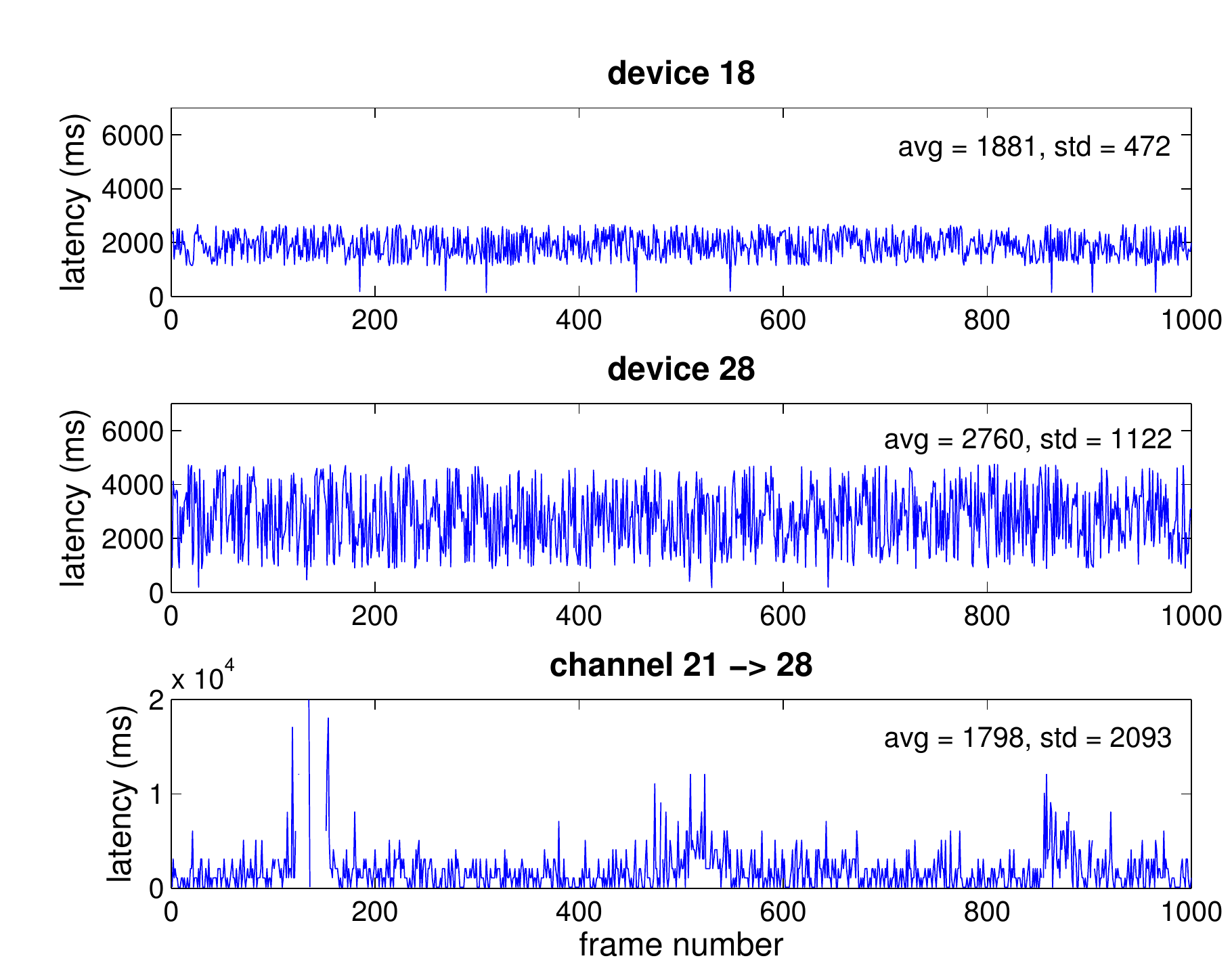}
\caption{Snapshots of measurement result: (a) device $18$'s computation latency (b) device $28$'s computation latency (c) transmission latency between them.}
\label{fig: samples}
\end{figure}

\begin{figure*}[!htb]
\minipage{0.32\textwidth}
  \centering
  \includegraphics[width=0.95\linewidth]{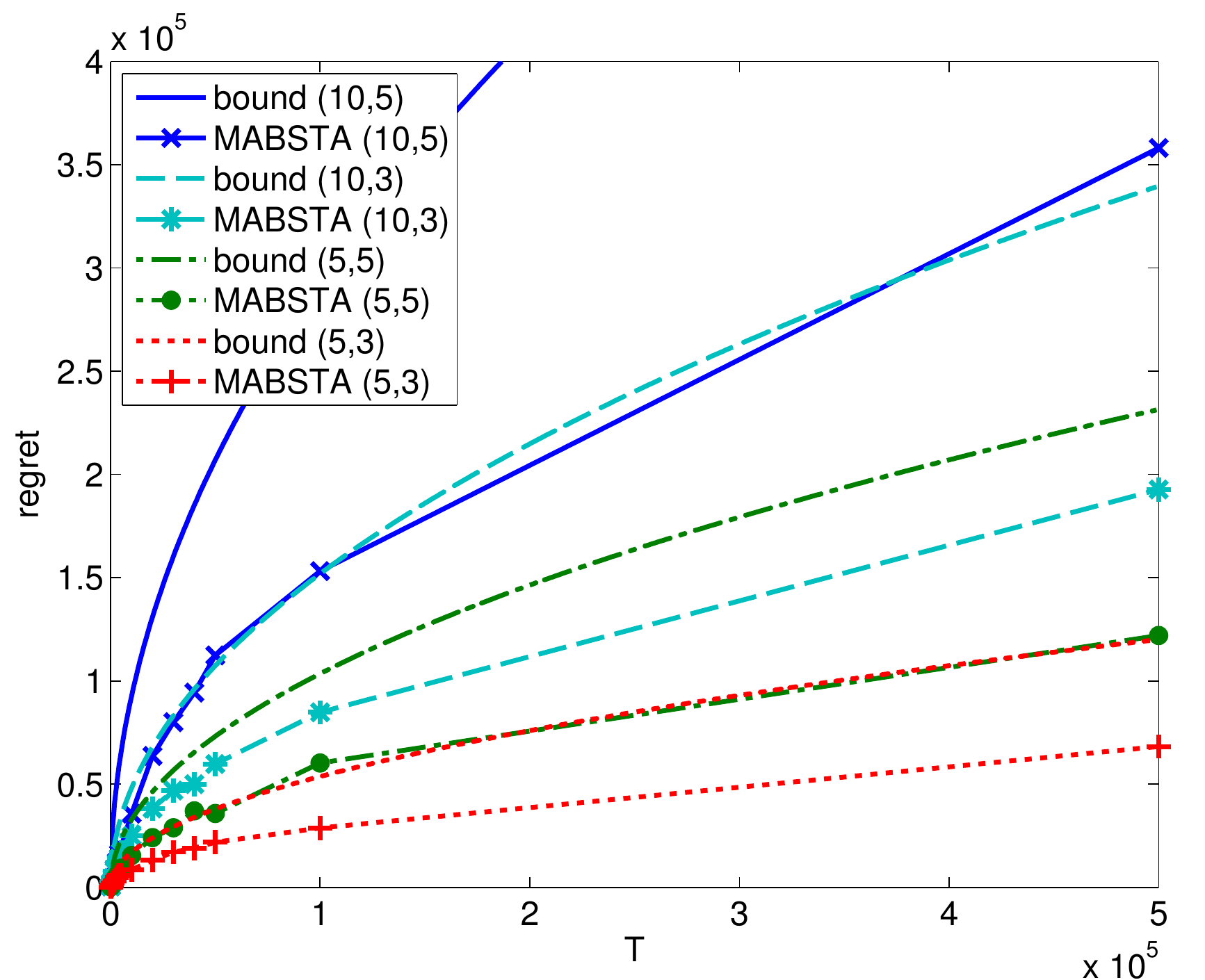}
  \caption{MABSTA's performance with upper bounds provided by Corollary \ref{col:1}}
  \label{fig: emulation_overall}
\endminipage\hfill
\minipage{0.32\textwidth}
  \centering
  \includegraphics[width=0.95\linewidth]{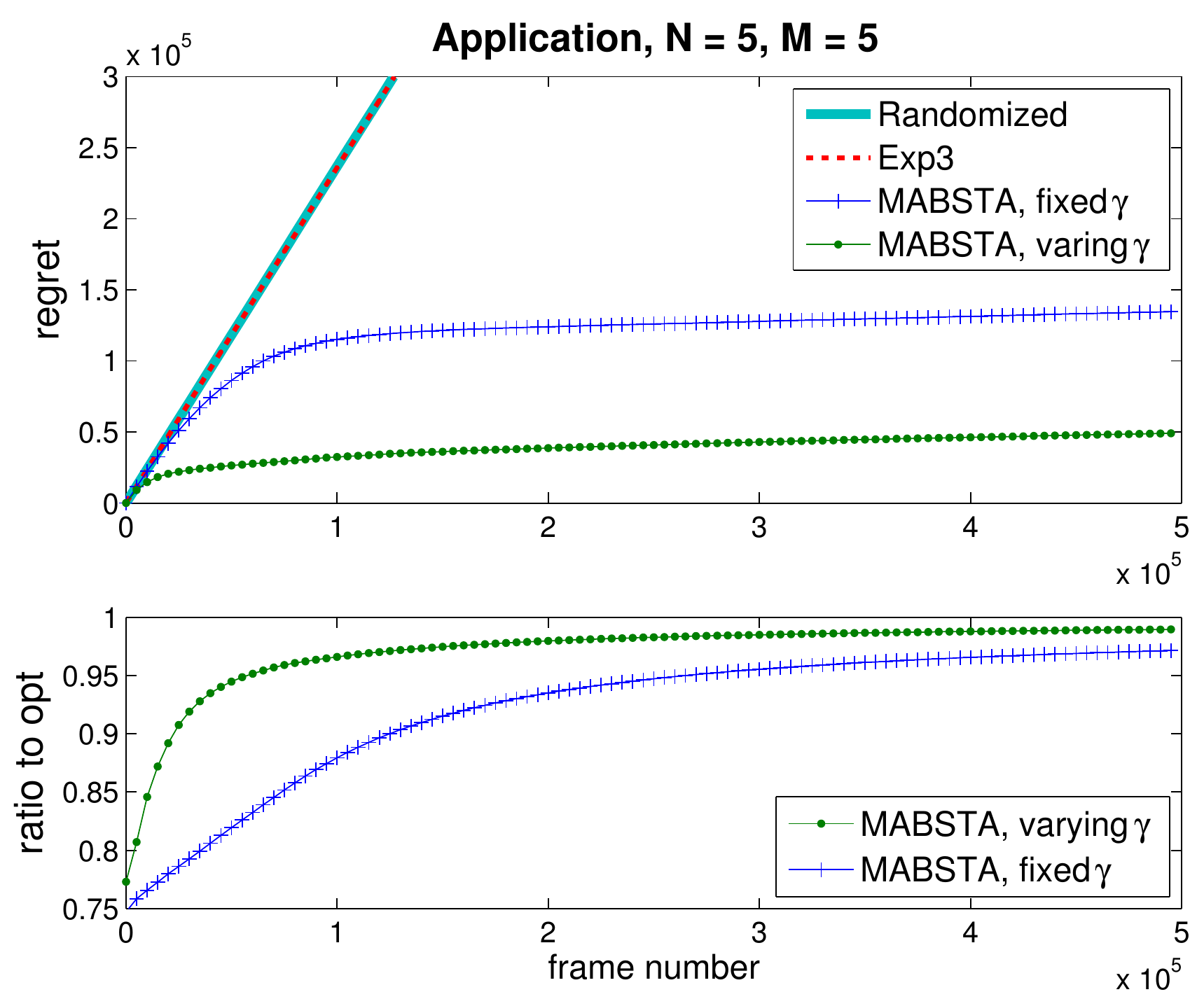}
  \caption{MABSTA compared with other algorithms for $5$-device network.}
  \label{fig: data_trace_N5_M5_with_ratio}
\endminipage\hfill
\minipage{0.32\textwidth}%
  \centering
  \includegraphics[width=0.95\linewidth]{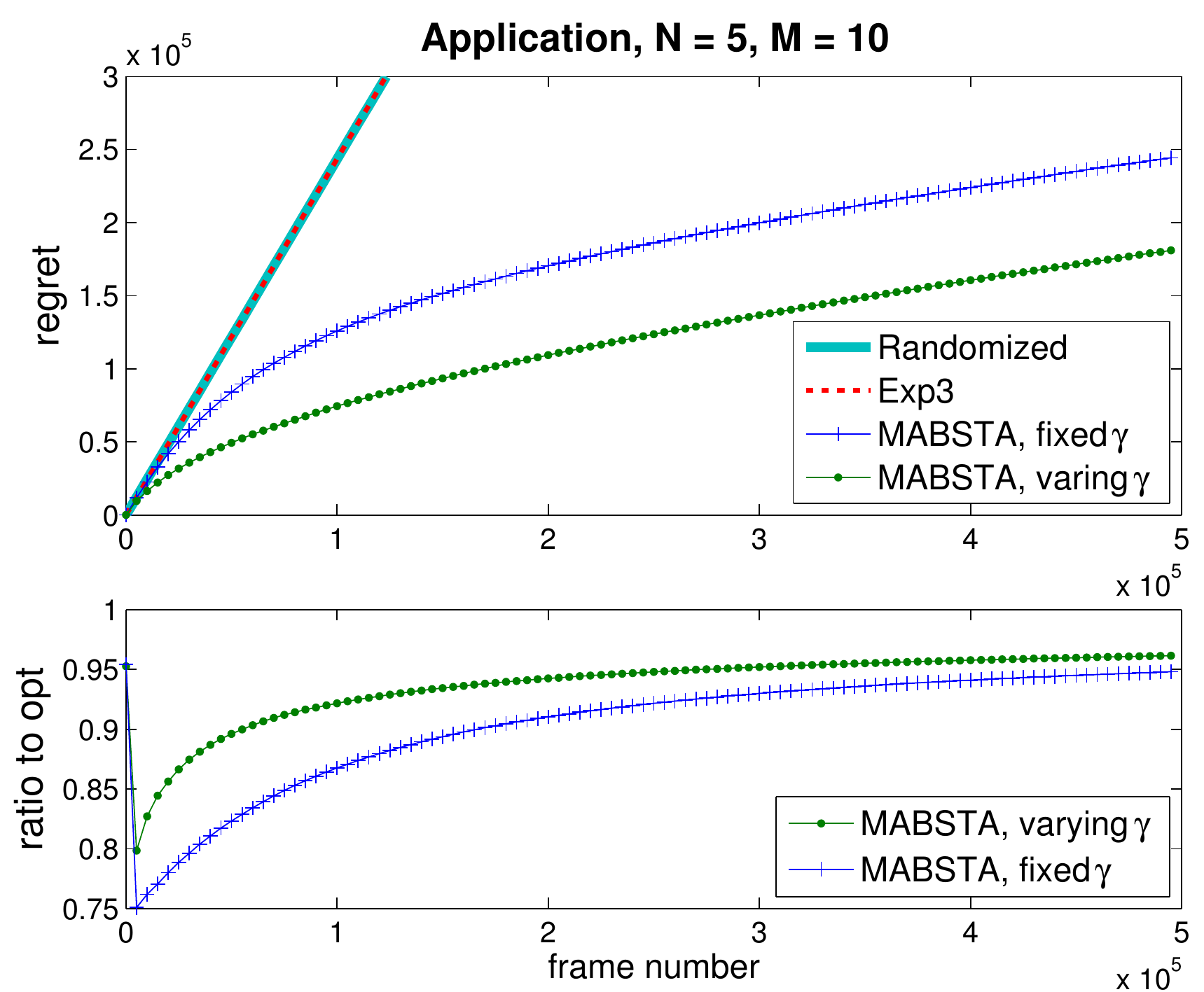}
  \caption{MABSTA compared with other algorithms for $10$-device network.}
\label{fig: data_trace_N5_M10_with_ratio}
\endminipage
\end{figure*}

To obtain trace data representative of a realistic environment, we run simulations on a large-scale wireless sensor network / IoT testbed. We create a network using 10 IEEE 802.15.4-based wireless embedded devices, and conduct a set of experiments to measure two performance characteristics utilized by MABSTA, namely channel conditions and computational resource availability. To assess the channel conditions, the time it takes to transfer $500$ bytes of data between every pair of motes is measured. To assess the resource availability of each device, we measure the amount of time it takes to run a simulated task for a uniformly distributed number of iterations. The parameters of the distribution are shown in Table \ref{table: parameter}. Since latency is positively correlated with device's energy consumption and the radio transmission power is kept constant in these experiments, it can also be used as an index for energy cost. We use these samples as the reward sequences in the following emulation.

We present our evaluation as the regret compared to the offline optimal solution in (\ref{eq:formulation}). For real applications the regret can be extra energy consumption over all nodes, or extra processing latency over all data frames. Figure \ref{fig: emulation_overall} validates MABSTA's performance guarantee for different problem sizes. From the cases we have considered, MABSTA's regret scales with $O(N^{1.5}M)$.

We further compare MABSTA with two other algorithms as shown in Figure \ref{fig: data_trace_N5_M5_with_ratio} and Figure \ref{fig: data_trace_N5_M10_with_ratio}. Exp3 is proposed for adversarial MAB in \cite{auer2002nonstochastic}. Randomized baseline simply selects an arm uniformly for each data frame. Applying Exp3 to our task assignment problem results in the learning time grows exponentially with $O(M^N)$. Hence, Exp3 is not competitive in our scheme, in which the regret grows nearly linear with $T$ as randomized baseline does. In addition to original MABSTA, we propose a more aggressive scheme by tuning $\gamma$ provided in MABSTA. That is, for each frame $t$, setting
\begin{equation}
\gamma_t = \min \left\{1,\sqrt{\frac{M(N+\left\vert{\mc{E}}\right\vert M) \ln M^N}{(e-1)(N+\left\vert{\mc{E}}\right\vert)t}}\right\}.
\label{eq: varying_gamma}
\end{equation}
From (\ref{eq: prob}), the larger the $\gamma$, the more chance that MABSTA will do exploration. Hence, by exploring more aggressively at the beginning and exploiting the best arm as $\gamma$ decreases with $t$, MABSTA with varying $\gamma$ learns the environment even faster and remains competitive with the offline optimal solution, where the ratio reaches $0.9$ at early stage. That is, after first $5000$ frames, MABSTA already achieves the performance at least $90\%$ of the optimal one. In sum, these empirical trace-based evaluations show that MABSTA scales well and outperforms the state of the art in adversarial online learning algorithms (EXP3). Moreover, it typically does significantly better in practice than the theoretical performance guarantee.

\section{Applications to Wireless Device Networks}
MABSTA is widely applicable to many realistic scenarios, including in the following device networks.
\subsection{Mobile Cloud Computing}
Computational offloading - migrating intensive tasks to more resourceful servers, has been a widely-used approach to augment computing on a resource-constrained device \cite{kumar2013_survey}. The performance of computational offloading on cellular networks varies with channel and server dynamics. Instead of solving deterministic optimization based on profiling, like MAUI \cite{cuervo2010_maui}, or providing a heuristic without performance guarantee, like Odessa \cite{ra2011odessa}, MABSTA can be applied to learn the optimal offloading decision (task assignment) in dynamic environment.

\subsection{Vehicular Ad Hoc Networks (VANETs)}
Applications on VANETs are acquiring commercial relevance recently. These applications, like content downloading, rely on both vehicle-to-vehicle (V2V) and vehicle-to-infrastructure (V2I) communications \cite{gerla2011vehicular}. Computational offloading, or service discovery over VANETs are promising approaches with the help by road-side units and other vehicles \cite{li2014computation}. How to leverage these intermittent connections and remote computational resources efficiently requires continuous run-time probing, which cannot be done by historical profiling due to fast-changing environment.

\subsection{Wireless Sensor Networks and IoT}
Wireless Sensor Networks (WSN) suffer from stringent energy usage on each node in real applications. These sensors are often equipped with functional microprocessors for some specific tasks. Hence, in some cases, WSN applications face the dilemma of pre-processing on less powerful devices or transmitting raw data to back-end processors \cite{viswanathan2013enabling}. Depending on channel conditions, MABTSA can adapt the strategies by assigning pre-processing tasks on front-end sensors when channel is bad, or simply forwarding raw data when channel is good. Moreover, MABSTA can also consider battery status so that the assignment strategy adapts to the battery remaining on each node in order to prolong network lifetime.

In the future IoT networks, fog computing is a concept similar to wireless distributed computing but scales to larger number of nodes and generalized heterogeneity on devices, communication protocols and deployment environment \cite{bonomi2012fog}. With available resources spread over the network, a high level programming model is necessary, where an interpreter takes care of task assignment and scheduling at run time \cite{hong2013mobile}. No single stochastic process can model this highly heterogeneous scheme. As an approach to stochastic online learning optimization, MABSTA provides a scalable approach and performance guarantee for this highly dynamic run-time environment.

\section{Conclusion}
With increasing number of devices capable of computing and communicating, the concept of Wireless Distributed Computing enables complex applications which a single device cannot support individually. However, the intermittent and heterogeneous connections and diverse device behavior make the performance highly-variant with time. In this paper, we have proposed a new online learning formulation for wireless distributed computing that does not make any stationary stochastic assumptions about channels and devices. We have presented MABSTA, which, to the best of our knowledge, is the first online learning algorithm tailored to this class of problems. We have proved that MABSTA can be implemented efficiently and provides performance guarantee for all dynamic environment. The trace-data emulation has shown that MABSTA is competitive to the optimal offline strategy and is adaptive to changes of the environment. Finally, we have identified several wireless distributed computing applications where MABSTA can be employed fruitfully.


%

\appendices
\section{Proof of Theorem \lowercase{\ref{th:1}}}

\begin{figure*}[!t]
\begin{align}
\notag \sum_{\mb{y} \in \mc{F}} p_{\mb{y}}(t)\hat{R}_{\mb{y}}(t)^2 &= \sum_{\mb{y} \in \mc{F}} p_{\mb{y}} \left( \sum_{i=1}^N\hat{R}_{i}^{(y_i)} + \sum_{(m,n) \in \mc{E}}\hat{R}_{mn}^{(y_my_n)}\right)^2 \\
&= \sum_{\mb{y} \in \mc{F}}p_{\mb{y}} \left(
\sum_{i,j} \hat{R}_{i}^{(y_i)}\hat{R}_{j}^{(y_j)}
+ \sum_{(m,n),(u,v)} \hat{R}_{mn}^{(y_my_n)}\hat{R}_{uv}^{(y_uy_v)} 
+ 2\sum_{i}\sum_{(m,n)}
\hat{R}_{i}^{(y_i)}\hat{R}_{mn}^{(y_my_n)}\right)
\label{eq:breakdown}
\end{align}
\hrule
\end{figure*}

We first prove the following lemmas. We will use more condensed notations like $\hat{R}_{i}^{(y_i)}$ for $\hat{R}_{i}^{(y_i)}(t)$ and $\hat{R}_{mn}^{(y_my_n)}$ for $\hat{R}_{mn}^{(y_my_n)}(t)$ in the prove where the result holds for each $t$.

\subsection{Proof of lemmas}
\begin{lemma}
\label{lemma: 1}
\vspace{-0.05in}
\begin{equation*}
\sum_{\mb{y} \in \mc{F}} p_{\mb{y}}(t)\hat{R}_{\mb{y}}(t) = \sum_{i=1}^N R_{i}^{(x^t_i)}(t) + \sum_{(m,n) \in \mc{E}}R_{mn}^{(x^t_mx^t_n)}(t).
\end{equation*}
\vspace{-0.05in}
\label{lm:1}
\end{lemma}
\begin{myproof}
\StartCompact{small}
\vspace{-0.05in}
\begin{align}
\notag \sum_{\mb{y} \in \mc{F}}p_{\mb{y}}(t)\hat{R}_{\mb{y}}(t) &= \sum_{\mb{y} \in \mc{F}}p_{\mb{y}} \left(\sum_{i=1}^N \hat{R}_{i}^{(y_i)} + \sum_{(m,n) \in \mc{E}}\hat{R}_{mn}^{(y_my_n)}\right) \\
&= \sum_i\sum_{\mb{y}}p_{\mb{y}}\hat{R}_{i}^{(y_i)} 
+ \sum_{(m,n)}\sum_{\mb{y}}p_{\mb{y}}\hat{R}_{mn}^{(y_my_n)},
\label{eq:lm1}
\end{align}
\vspace{-0.05in}
\StopCompact{small}
where
\vspace{-0.05in}
\[
\sum_{\mb{y}}p_{\mb{y}}\hat{R}_{i}^{(y_i)} = \sum_{\mb{y} \in \mc{C}_{ex}^i} p_{\mb{y}} \frac{R_{i}^{(x^t_i)}}{\sum_{\mb{z} \in \mc{C}_{ex}^i}p_{\mb{z}}} = R_{i}^{(x^t_i)},
\]
\vspace{-0.05in}
and similarly,
\vspace{-0.05in}
\[
\sum_{\mb{y}}p_{\mb{y}}\hat{R}_{mn}^{(y_my_n)} = R_{mn}^{(x^t_mx^t_n)}.
\]
\vspace{-0.05in}
Applying the result to (\ref{eq:lm1}) completes the proof.
\end{myproof}

\begin{lemma}
\label{lemma: 2}
For all $\mb{y} \in \mc{F}$, we have
\vspace{-0.05in}
\begin{equation*}
\mbb{E}\{\hat{R}_{\mb{y}}(t)\} = \sum_{i=1}^N R_{i}^{(y_i)}(t) + \sum_{(m,n) \in \mc{E}}R_{mn}^{(y_my_n)}(t).
\end{equation*}
\vspace{-0.05in}
\end{lemma}
\begin{myproof}
\vspace{-0.05in}
\begin{equation}
\mbb{E}\{\hat{R}_{\mb{y}}(t)\} = \sum_{i=1}^N \mbb{E}\{\hat{R}_{i}^{(y_i)}\} + \sum_{(m,n) \in \mc{E}} \mbb{E}\{\hat{R}_{mn}^{(y_my_n)}\},
\end{equation}
\vspace{-0.05in}
where
\vspace{-0.05in}
\[
\mbb{E}\{\hat{R}_{i}^{(y_i)}\} = \mbb{P}\{x^t_i = y_i\}\frac{R_{i}^{(y_i)}}{\sum_{\mb{z} \in \mc{C}_{ex}^i}p_{\mb{z}}} = R_{i}^{(y_i)},
\]
and similarly, $\mbb{E}\{\hat{R}_{mn}^{(y_my_n)}\} = R_{mn}^{(y_my_n)}$.
\end{myproof}

\begin{lemma}
\label{lemma: 3}
If $\mc{F} = \{\mb{x} \in [M]^N\}$, then for $M \geq 3$ and $\E \geq 3$,
\begin{equation*}
\sum_{\mb{y} \in \mc{F}} p_{\mb{y}}(t)\hat{R}_{\mb{y}}(t)^2 \leq \frac{\E}{M^{N-2}} \sum_{\mb{y} \in \mc{F}} \hat{R}_{\mb{y}}(t).
\end{equation*}
\end{lemma}
\begin{myproof}
We first expand the left-hand-side of the inequality as shown in (\ref{eq:breakdown}) at the top of this page. In the following, we derive the upper bound for each term in (\ref{eq:breakdown}) for all $i \in [N]$, $(m,n) \in \mc{E}$.
\StartCompact{small}
\begin{align}
\notag \sum_{\mb{y}} &p_{\mb{y}}\hat{R}_{i}^{(y_i)}\hat{R}_{j}^{(y_j)} = \sum_{\mb{y} \in \mc{C}_{ex}^i \cap \mc{C}_{ex}^j} p_{\mb{y}}\frac{R_{i}^{(x^t_i)}R_{j}^{(x^t_j)}}{\sum_{\mb{z}\in \mc{C}_{ex}^i}p_{\mb{z}} \cdot \sum_{\mb{z}\in \mc{C}_{ex}^j}p_{\mb{z}}} \\
& \leq R_{j}^{(x^t_j)} \frac{R_{i}^{(x^t_i)}}{\sum_{\mb{z}\in \mc{C}_{ex}^i}p_{\mb{z}}} 
= R_{j}^{(x^t_j)} \hat{R}_{i}^{(x^t_i)}
\leq \frac{1}{M^{N-1}}\sum_{\mb{y}}\hat{R}_{i}^{(y_i)}
\label{eq:1term}
\end{align}
\StopCompact{small}
The first inequality in (\ref{eq:1term}) follows by $\mc{C}_{ex}^i \cap \mc{C}_{ex}^j$ is a subset of $\mc{C}_{ex}^j$ and the last inequality follows by $\hat{R}_{i}^{(y_i)} = \hat{R}_{i}^{(x^t_i)}$ for all $\mb{y}$ in $\mc{C}_{ex}^i$. Hence,
\begin{equation}
\sum_{i,j}\sum_{\mb{y}}p_{\mb{y}}\hat{R}_{i}^{(y_i)}\hat{R}_{j}^{(y_j)} \leq  \frac{1}{M^{N-2}}\sum_{\mb{y}}\sum_{i}\hat{R}_{i}^{(y_i)}.
\label{eq:1term_result}
\end{equation}
Similarly,
\begin{equation}
\sum_{(m,n),(u,v)}\sum_{\mb{y}}p_{\mb{y}}\hat{R}_{mn}^{(y_my_n)}\hat{R}_{uv}^{(y_uy_v)} \leq \frac{\E}{M^{N-2}}\sum_{\mb{y}}\sum_{(m,n)}\hat{R}_{mn}^{(y_my_n)}.
\label{eq:2term_result}
\end{equation}
For the last term in (\ref{eq:breakdown}), following the similar argument gives
\StartCompact{small}
\begin{align*}
\sum_{\mb{y}} & p_{\mb{y}}\hat{R}_{i}^{(y_i)}\hat{R}_{mn}^{(y_my_n)} = \sum_{\mb{y} \in \mc{C}_{ex}^i \cap \mc{C}_{tx}^{mn}} p_{\mb{y}}\frac{R_{i}^{(x^t_i)}R_{mn}^{(x^t_mx^t_n)}}{\sum_{\mb{z}\in \mc{C}_{ex}^i}p_{\mb{z}} \cdot \sum_{\mb{z}\in \mc{C}_{tx}^{mn}}p_{\mb{z}}} \\
& \leq R_{mn}^{(x^t_mx^t_n)} \frac{R_{i}^{(x^t_i)}}{\sum_{\mb{z}\in \mc{C}_{ex}^i}p_{\mb{z}}} = R_{mn}^{(x^t_mx^t_n)} \hat{R}_{i}^{(x^t_i)} \leq \frac{1}{M^{N-1}}\sum_{\mb{y}}\hat{R}_{i}^{(y_i)}.
\end{align*}
\StopCompact{small}
Hence,
\begin{equation}
\sum_{i}\sum_{(m,n)}\sum_{\mb{y}}p_{\mb{y}}\hat{R}_{i}^{(y_i)}\hat{R}_{mn}^{(y_my_n)} \leq  \frac{\E}{M^{N-1}}\sum_{\mb{y}}\sum_{i}\hat{R}_{i}^{(y_i)}.
\label{eq:3term_result}
\end{equation}
Applying (\ref{eq:1term_result}), (\ref{eq:2term_result}) and (\ref{eq:3term_result}) to (\ref{eq:breakdown}) gives
\StartCompact{small}
\begin{align}
\notag \sum_{\mb{y} \in \mc{F}} &p_{\mb{y}}(t)\hat{R}_{\mb{y}}(t)^2 \\
\notag &\leq \sum_{\mb{y} \in \mc{F}} [\sum_{i} (\frac{1}{M^{N-2}}+\frac{2\E}{M^{N-1}})\hat{R}_{i}^{(y_i)} + \sum_{(m,n)}\frac{\E}{M^{N-2}}\hat{R}_{mn}^{(y_my_n)} ] \\
&\leq \frac{\E}{M^{N-2}}\sum_{\mb{y} \in \mc{F}}\hat{R}_{\mb{y}}(t).
\label{eq:lm3}
\end{align}
\StopCompact{small}
The last inequality follows by the fact that $\frac{1}{M^{N-2}}+\frac{2\E}{M^{N-1}} \leq \frac{\E}{M^{N-2}}$ for $M \geq 3$ and $\E \geq 3$. For $M = 2$, we have
\[
\sum_{\mb{y} \in \mc{F}} p_{\mb{y}}(t)\hat{R}_{\mb{y}}(t)^2 \leq \frac{M+2\E}{M^{N-1}}\sum_{\mb{y} \in \mc{F}}\hat{R}_{\mb{y}}(t). 
\]
Since we are interested in the regime where (\ref{eq:lm3}) holds, we will use this result in our proof of Theorem \ref{th:1}.
\end{myproof}

\begin{lemma}
\label{lemma: 4}
Let $\alpha = \frac{\gamma}{M(N+\E M)}$, if $\mc{F} = \{\mb{x} \in [M]^N\}$, then for all $\mb{y} \in \mc{F}$, all $t = 1,\cdots,T$, we have $\alpha \hat{R}_{\mb{y}}(t) \leq 1$.
\end{lemma}
\begin{myproof}
Since $\left\vert\mc{C}_{ex}^i\right\vert \geq M^{N-1}$ and $\left\vert\mc{C}_{tx}^{mn}\right\vert \geq M^{N-2}$ for all $i \in [N]$ and $(m,n) \in \mc{E}$, each term in $\hat{R}_{\mb{y}}(t)$ can be upper bounded as
\begin{align}
\hat{R}_{i}^{(y_i)} &\leq \frac{R_{i}^{(y_i)}}{\sum_{\mb{z} \in \mc{C}_{ex}^i}p_{\mb{z}}} \leq \frac{1}{M^{N-1}\frac{\gamma}{M^{N}}} = \frac{M}{\gamma}, \\
\hat{R}_{i}^{(y_{i-1}y_i)} &\leq \frac{R_{i}^{(y_{i-1}y_i)}}{\sum_{\mb{z} \in \mc{C}_{tx}^i}p_{\mb{z}}} \leq \frac{1}{M^{N-2}\frac{\gamma}{M^{N}}} = \frac{M^2}{\gamma}.
\end{align}
Hence, we have
\begin{align}
\notag \hat{R}_{\mb{y}}(t) &= \sum_{i=1}^N \hat{R}_{i}^{(y_i)} + \sum_{(m,n)\in \mc{E}} \hat{R}_{mn}^{(y_my_n)}\\
& \leq N\frac{M}{\gamma} + \E\frac{M^2}{\gamma} = \frac{M}{\gamma}(N+\E M).
\end{align}
Let $\alpha = \frac{\gamma}{M(N+ \E M)}$, we achieve the result.
\end{myproof}

\subsection{Proof of Theorem \lowercase{\ref{th:1}}}
\begin{myproof}
Let $W_t = \sum_{\mb{y} \in \mc{F}} w_{\mb{y}}(t)$. We denote the sequence of decisions drawn at each frame as $\mb{x}=[\mb{x}^1,\cdots,\mb{x}^T]$, where $\mb{x}^t \in \mc{F}$ denotes the arm drawn at step $t$. Then for all data frame $t$,
\StartCompact{small}
\begin{align}
\notag \frac{W_{t+1}}{W_t} =& \sum_{\mb{y} \in \mc{F}}\frac{w_{\mb{y}}(t)}{W_t}\exp\left(\alpha \hat{R}_{(\mb{y})}(t)\right) \\
\notag =& \sum_{\mb{y} \in \mc{F}}\frac{p_{\mb{y}}(t)-\tfrac{\gamma}{\left\vert\mc{F}\right\vert}}{1-\gamma}\exp\left(\alpha \hat{R}_{(\mb{y})}(t)\right) \\
\leq& \sum_{\mb{y} \in \mc{F}}\frac{p_{\mb{y}}(t)-\tfrac{\gamma}{\left\vert\mc{F}\right\vert}}{1-\gamma} \left(1 + \alpha \hat{R}_{(\mb{y})}(t) + (e-2) \alpha^2 \hat{R}_{(\mb{y})}(t)^2\right) \label{eq:upperBound} \\
\notag \leq& 1 + \frac{\alpha}{1-\gamma}\left(\sum_{i=1}^N R^{(x^t_i)}_{i}(t) + \sum_{(m,n) \in \mc{E}}R^{(x^t_mx^t_n)}_{mn}(t)\right) \\ 
&+ \frac{(e-2)\alpha^2}{1-\gamma}\frac{\E}{M^{N-2}}\sum_{\mb{y} \in \mc{F}} \hat{R}_{\mb{y}}(t).
\label{eq:upperBound2}
\end{align}
\StopCompact{small}
Eq. (\ref{eq:upperBound}) follows by the fact that $e^x \leq 1+x+(e-2)x^2$ for $x \leq 1$. Applying Lemma \ref{lemma: 1} and Lemma \ref{lemma: 3} we arrive at (\ref{eq:upperBound2}). Using $1+x \leq e^x$ and taking logarithms at both sides,
\StartCompact{small}
\begin{align}
\notag \ln \frac{W_{t+1}}{W_t} \leq & \frac{\alpha}{1-\gamma}\left(\sum_{i=1}^N R^{(x^t_i)}_{i}(t) + \sum_{(m,n) \in \mc{E}}R^{(x^t_mx^t_n)}_{mn}(t)\right) \\ 
\notag &+ \frac{(e-2)\alpha^2}{1-\gamma}\frac{\E}{M^{N-2}}\sum_{\mb{y} \in \mc{F}} \hat{R}_{\mb{y}}(t).
\end{align}
\StopCompact{small}
Taking summation from $t=1$ to $T$ gives
\StartCompact{small}
\vspace{-0.05in}
\begin{equation}
\ln \frac{W_{T+1}}{W_1} \leq \frac{\alpha}{1-\gamma}\hat{R}_{total} + \frac{(e-2)\alpha^2}{1-\gamma}\frac{\E}{M^{N-2}}\sum_{t=1}^{T}\sum_{\mb{y} \in \mc{F}} \hat{R}_{\mb{y}}(t).
\label{eq:upBd}
\end{equation}
\StopCompact{small}
On the other hand,
\StartCompact{small}
\vspace{-0.05in}
\begin{equation}
\ln \frac{W_{T+1}}{W_1} \geq \ln \frac{w_{\mb{z}}(T+1)}{W_1} = \alpha \sum_{t=1}^{T}\hat{R}_{\mb{z}}(t)-\ln M^N, \; \forall \mb{z} \in \mc{F}.
\label{eq:lowBd}
\end{equation}
\StopCompact{small}
Combining (\ref{eq:upBd}) and (\ref{eq:lowBd}) gives
\StartCompact{small}
\vspace{-0.05in}
\begin{equation}
\hat{R}_{total} \geq (1-\gamma)\sum_{t=1}^{T}\hat{R}_{\mb{z}}(t)- (e-2)\alpha \frac{\E}{M^{N-2}}\sum_{t=1}^{T}\sum_{\mb{y} \in \mc{F}}\hat{R}_{\mb{y}}(t) - \frac{\ln M^N}{\alpha}.
\label{eq:sandwich}
\end{equation}
\StopCompact{small}
Eq. (\ref{eq:sandwich}) holds for all $\mb{z} \in \mc{F}$. Choose $\mb{x}^\star$ to be the assignment strategy that maximizes the objective in (\ref{eq:formulation}). Now we take expectations on both sides based on $\mb{x}^1,\cdots,\mb{x}^T$ and use Lemma \ref{lemma: 2}. That is,
\StartCompact{small}
\vspace{-0.05in}
\begin{equation*}
\sum_{t=1}^{T}\mbb{E}\{\hat{R}_{\mb{x}^\star}(t)\} = \sum_{t=1}^{T}[\sum_{i=1}^N R_{i}^{(x^\star_i)}(i) + \sum_{(m,n) \in \mc{E}} R_{mn}^{(x^\star_m x^\star_n)}(t)] = R_{total}^{max},
\end{equation*}
\StopCompact{small}
and
\StartCompact{small}
\begin{align*}
\sum_{t=1}^{T}&\sum_{\mb{y} \in \mc{F}}\mbb{E}\{\hat{R}_{\mb{y}}(t)\} \\
&= \sum_{t=1}^{T}\sum_{\mb{y} \in \mc{F}}\left(\sum_{i=1}^N R_{i}^{(y_i)}(t) + \sum_{(m,n) \in \mc{E}} R_{mn}^{(y_my_n)}(t)\right) \leq M^NR_{total}^{max}.
\end{align*}
\StopCompact{small}
Applying the result to (\ref{eq:sandwich}) gives
\StartCompact{small}
\begin{equation*}
\mbb{E}\{\hat{R}_{total}\} \geq (1-\gamma)R_{total}^{max} - \E M^2(e-2)\alpha R_{total}^{max} - \frac{\ln M^N}{\alpha}.
\end{equation*}
\StopCompact{small}
Let $\alpha = \frac{\gamma}{M(N+\E M)}$, we arrive at
\StartCompact{small}
\begin{equation*}
R_{total}^{max} - \mbb{E}\{\hat{R}_{total}\} \leq (e-1)\gamma R_{total}^{max} + \frac{M(N+\E M) \ln M^N}{\gamma}.
\end{equation*}
\StopCompact{small}
\end{myproof}

%
\bibliographystyle{IEEEtran}
\bibliography{IEEEabrv,main}

%

\end{document}